\documentclass[letterpaper, 10 pt, conference]{ieeeconf}  

\IEEEoverridecommandlockouts                              

\usepackage{amsmath} 
\usepackage{amssymb}
\usepackage{bm}
\usepackage{booktabs}  
\usepackage{graphicx}
\usepackage{caption}
\usepackage{xcolor}
\usepackage{hyperref}
\usepackage[table]{xcolor}

\title{\LARGE \bf
PGMT: Perceptive General Motion Tracking for Humanoid Robots
}

\author{Hongyi LI$^{1}$, LI Peizhuo$^{2}$, Yucheng TAO$^{1}$, Ze WANG$^{1}$, Fangzhou XU$^{2}$, Jinyi CHEN$^{1}$, Yanyan YUAN$^{3}$, \\
Dapeng JIA$^{2}$, Yongbin JIN$^{3,\dagger}$, Mingfeng FAN$^{2,\dagger}$, Guillaume SARTORETTI$^{2}$ and Hongtao WANG$^{1,3,\ddagger}$
\thanks{$^{\dagger}$Corresponding Author \hspace{10pt} $^{\ddagger}$Project Lead}
\thanks{$^{1}$Center of X-Mechanics, Zhejiang University, Hangzhou, China.}
\thanks{$^{2}$MARMot Lab, National University of Singapore, Singapore}
\thanks{$^{3}$MirrorMe Technology Co., Ltd.}
}

\begin{document}

\IEEEaftertitletext{%
  \noindent\begin{minipage}{\textwidth}
    \centering
    \includegraphics[width=1\textwidth]{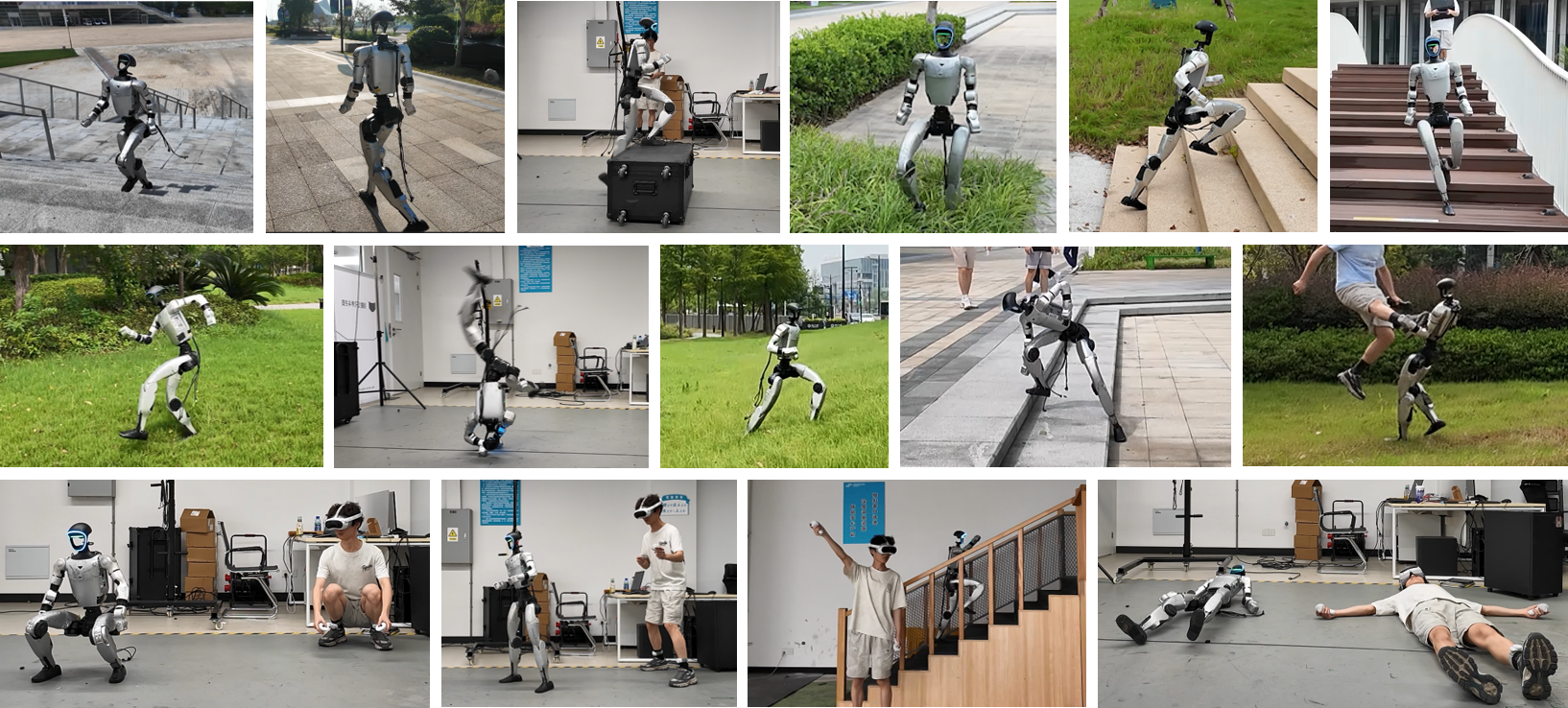}
    \captionof{figure}{\textbf{PGMT robustly executes whole-body motions across diverse real-world environments}, from structured indoor settings to stairs, uneven outdoor terrain, and dense vegetation. The same policy accommodates motion references from multiple command sources, providing a unified interface for perceptive whole-body locomotion and tracking.}
    \label{fig:frontpage}
    \vspace{5pt}
  \end{minipage}%
}
\maketitle
\thispagestyle{empty}
\pagestyle{empty}

\begin{abstract}

Humanoid motion trackers can reproduce diverse whole-body motions, but their performance degrades on complex terrain where terrain-agnostic references become physically infeasible. We present \textbf{PGMT}, a \underline{P}erceptive \underline{G}eneral \underline{M}otion \underline{T}racking pipeline for humanoid robots that learns terrain adaptation from independently selected motion references and terrains. PGMT first learns a general tracking and recovery prior, then incorporates terrain perception through motion-conditioned terrain glimpses that selectively encode regions relevant to the current motion. Terrain-aware tracking relaxation allows necessary deviations from the reference while preserving its motion intent. Zero-shot deployment on a Unitree G1 demonstrates robust terrain-adaptive locomotion and whole-body motion execution over real-world terrain with obstacles up to 37 cm high, while supporting teleoperation, dynamic motion tracking, and fall recovery. PGMT extends general humanoid motion tracking beyond flat ground, providing a unified policy for terrain-adaptive locomotion, diverse whole-body behaviors, and teleoperation in complex environments. Project homepage: \url{https://luyili.github.io/pgmt/}

\end{abstract}

\section{Introduction}

Recent advances in humanoid control have produced highly capable whole-body motion trackers that can reproduce diverse human motions with remarkable fidelity~\cite{he2024learning,he2025hover}. Yet their success remains largely confined to flat ground. In complex environments, terrain-agnostic references may become physically infeasible, requiring the robot to actively adapt its contacts, swing trajectories, and whole-body posture according to local terrain observations rather than strictly reproduce the commanded motion~\cite{wang2026perceptive,zhang2026meshmimic}.

Such perceptive adaptation is particularly important for applications such as remote teleoperation and data collection, search and rescue over rough terrain, and transportation through complex environments~\cite{fu2024humanplus}. In these scenarios, motion commands from human operators or higher-level planners cannot be expected to anticipate the exact geometry encountered by the robot. A general motion tracker should therefore use onboard perception to resolve local terrain constraints autonomously: the reference specifies \emph{what motion to perform}, while environmental observations determine \emph{how that motion should be physically realized}.

Learning such a capability, however, is difficult because large-scale motion and terrain data are naturally unpaired. Existing human motion datasets provide diverse whole-body behaviors but generally lack synchronized terrain geometry and terrain-specific adaptations~\cite{mahmood2019amass}, while collecting paired motion--terrain demonstrations becomes increasingly difficult as motion and terrain diversity grow. Recent work addresses this problem by constructing terrain-adapted references before tracking \cite{zhang2026learning}. Although effective, these approaches depend on the quality and generalization of the intermediate stage, while the need to construct terrain-adapted references limits their scalability to large-scale motion and terrain data. A general perceptive tracker should instead learn to relate motion intent directly to terrain constraints from independently available motion and terrain data.

We instead treat terrain adaptation as a capability of the tracking policy itself. The raw reference specifies the intended motion, while terrain perception provides the local physical constraints. By jointly processing both signals at the same level in an Intent Fusion Module (IFM), the policy can deviate from the reference when necessary to maintain feasibility, without requiring an upstream terrain-matched reference generator. Building on this formulation, we propose \emph{PGMT}, a Perceptive General Motion Tracking pipeline that takes raw reference motions and local terrain observations directly as input. PGMT learns to attend to terrain regions relevant to the current motion and to adjust footholds and whole-body posture when the environment makes strict reference tracking infeasible, while preserving the underlying motion intent.

This formulation turns perceptive tracking into a reusable low-level interface between terrain-agnostic motion commands and terrain-aware physical execution. The same PGMT policy can execute dynamic whole-body motions over stairs, slopes, and uneven ground, serve as a perceptive locomotion controller, and support teleoperation in which the robot resolves local terrain constraints on the operator's behalf. More broadly, motion references from human operators, motion-matching systems, or higher-level planners can specify desired behavior without explicitly encoding the geometry under the robot's feet, while PGMT grounds these commands into physically feasible whole-body motion. The contributions of this work are as follows.

\begin{itemize}
\item We propose \textbf{PGMT}, a general motion tracker with perception that directly tracks terrain-agnostic motion references over complex terrain without requiring terrain-matched reference trajectories.
\item We design a \emph{progressively extensible architecture} that decouples motion and perception encoding, enabling new perceptual modalities to be injected while preserving the learned motion tracking capability.
\item We conduct extensive simulations and real-world experiments across diverse terrains and motion-reference command sources, demonstrating the robust zero-shot generalization of PGMT.

\end{itemize}

\begin{figure*}
    \centering
    \includegraphics[width=1\linewidth]{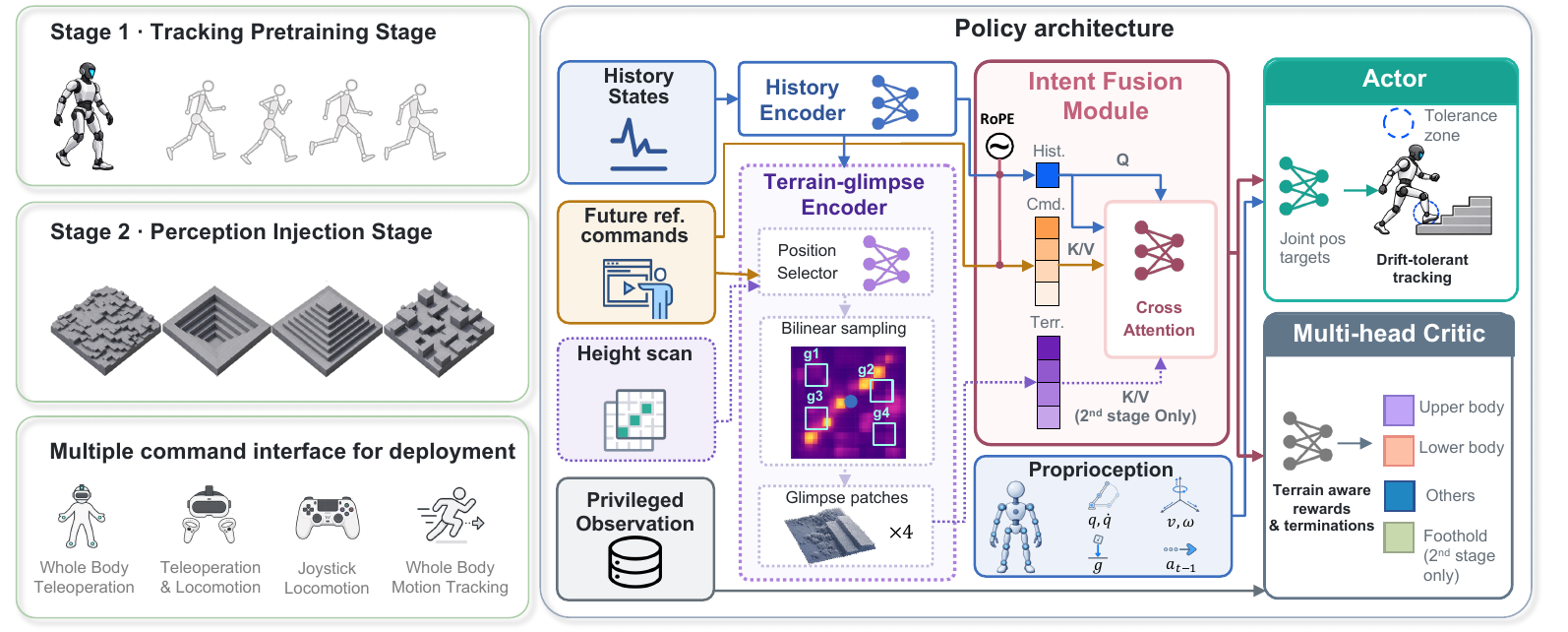}
    \caption{\textbf{Overview of PGMT.} PGMT is trained in two stages: a Tracking Pretraining Stage that establishes a general whole-body motion tracking and recovery prior, followed by a Perception Injection Stage that selectively fuses local geometry with motion intent, enabling terrain adaptation without terrain-matched reference trajectories. The resulting unified policy supports terrain-adaptive locomotion, whole-body motion tracking, and multiple teleoperation interfaces.}
    \label{fig:pipeline}
    \vspace{-20pt}
\end{figure*}

\section{Related Works}

\noindent\textbf{Large-Scale Motion Priors for Humanoid Control.} General humanoid motion tracking grew out of large-scale retargeting and privileged policy distillation, enabling expressive real-world control through flexible command interfaces~\cite{he2025hover,he2024omnih2o,ji2412exbody2}. As the motion repertoire expanded, adaptive sampling, specialized experts, and latent representations addressed uneven motion difficulty and limited policy capacity~\cite{chen2025gmt,yin2026unitracker}. Embodiment mismatch can be handled by adapting motion before policy learning or transferring a tracker after training~\cite{zhao2025smap,yang2026any2any}. Recovery training and dynamics adaptation instead target failures during deployment~\cite{ma2026robust,he2025asap}. Scaling data and compute broadens behavior coverage, while masked conditioning supports varied command formats~\cite{luo2026sonic, tessler2024maskedmimic}. Even so, tracking depends on reference quality because retargeting artifacts and infeasible poses are passed to the policy~\cite{araujo2025retargeting}. The references also lack the geometry encountered at deployment and may conflict with terrain or contact constraints.

\noindent\textbf{Perception-Aware Humanoid Control.} Perception changes the problem from reproducing a reference to deciding how it should be executed on the observed terrain. Early learning-based approaches demonstrated whole-body obstacle crossing and used teacher--student training to cope with noisy terrain estimates~\cite{zhuang2024humanoid,sun2025learning}. Later work focuses on both the reliability and relevance of perceptual input. Depth synthesis and global--local reasoning improve terrain representations~\cite{sun2026dpl,fu2026global}, whereas elevation maps and active gaze focus control on regions that matter for the current motion~\cite{wang2025beamdojo,li2026taga}. In broader legged locomotion, separate proprioceptive and visuospatial pathways allow stable control when vision fails~\cite{li2025kivi}, and perception has also been connected with longer-horizon navigation~\cite{han2026guidewalk}. These methods adapt well to terrain, but their objectives remain tied to particular locomotion or obstacle-crossing tasks rather than diverse whole-body references.

\noindent\textbf{Perceptive Motion Priors and Tracking.}
Combining motion priors with perception exposes a data problem: large motion collections are rarely paired with terrain-specific adaptations. Existing work differs mainly in where it adds this missing information. Some methods construct supervision before policy learning from terrain-conformal references, scene geometry, motion-matched experts, or inferred contacts~\cite{wang2026perceptive,zhang2026meshmimic,wu2026perceptive,chen2026scenebot}. Others adapt within a generation-tracking loop, using execution feedback to reshape references or fine-tune the tracker~\cite{zhang2026learning,ling2026gentrack}. Guided diffusion provides another route by building behaviors from motion primitives collected by a real-robot tracker~\cite{liao2025beyondmimic}. These strategies avoid direct motion--terrain pairing, but still rely on intermediate signals whose errors propagate to the tracker.

\section{Problem Formulation}
\label{sec:problem}

We formulate perceptive whole-body motion tracking as a partially observable Markov decision process (POMDP). At each time step $t$, the policy $p_\theta$ takes an observation as input and outputs the target joint positions $\bm{a_t}\in \mathbb{R}^{29}$ as the action. The full perceptive observation is given by $\bm{y}_t=\{\bm{o}_t, \bm{H}_t, \bm{C}_t^{K}, \bm{M}_t\} $, where $\bm{o}_t\in\mathbb{R}^{96}$ denotes the current proprioception for immediate feedback, $\bm{H}_t\in\mathbb{R}^{10\times96}$ represents a ten-frame proprioception history, $\bm{C}_t^{K}$ denotes future reference frames, and $\bm{M_t}$ represents the elevation map. Each frame in $\bm{o}_t$ and $\bm{H}_t$ contains reference-relative anchor orientation $\bm{e}_{t}\in \mathbb{R}^{6}$, base angular velocity ${\bm{\omega}}_{t}\in \mathbb{R}^{3}$, joint positions ${\bm{q}}_{t}\in \mathbb{R}^{29}$ and velocities $\dot{\bm{q}}_{t}\in \mathbb{R}^{29}$, and the previous action $\bm{a}_{t-1}$. We sample the future reference frames at exponentially increasing temporal offsets:
\begin{equation}
  \begin{aligned}
    \bm{C}_t^K&=([\bm{q}^{r}_{\tau},\dot{\bm{q}}^{r}_{\tau},
      \widetilde{\bm{v}}^{r}_{\tau}])_{k=0}^{K-1}
      \in\mathbb{R}^{K\times61},\\
  \end{aligned}
    \label{eq:reference}
\end{equation}
where $k$ indexes the future frames and $\tau=t+2^k-1$. The superscript $r$ denotes reference quantities. Specifically, $\bm{q}^{r}\in \mathbb{R}^{29}$ and $\dot{\bm{q}}^{r}\in \mathbb{R}^{29}$ denote reference joint positions and velocities, respectively, while $\widetilde{\bm{v}}^{r}=(\widetilde v_x,\widetilde v_y,0)$ denotes the corrected planar anchor-velocity command. Collectively, $\bm{C}_t^K$ provides the future motion intent to the policy. The elevation map $\bm{M}_t\in\mathbb{R}^{21\times21}$ covers a yaw-aligned $2\,\mathrm{m}\times2\,\mathrm{m}$ region and is introduced during the second training stage.

\section{Perceptive Generalized Motion Tracking}

As illustrated in Fig.~\ref{fig:pipeline}, PGMT is trained in two stages. The \emph{Tracking Pretraining Stage} first learns a general terrain-agnostic prior for whole-body motion tracking, while the \emph{Perception Injection Stage} subsequently incorporates terrain perception to enable terrain-aware adaptation without relearning the underlying tracking capability from scratch. In the first stage, the policy $p_\theta$ receives the non-perceptive observation $\bm{y}_t^{pre} =\{\bm{o}_t,\bm{H}_t,\bm{C}_t^K\}$ as input. In the second stage, the elevation map $\bm{M}_t$ is additionally introduced, yielding the full perceptive observation $\bm{y}_t^{\mathrm{inj}}=\{\bm{o}_t,\bm{H}_t,\bm{C}_t^K,\bm{M}_t\}$.

\subsection{Tracking Pretraining Stage}
\label{sec:pretraining}

The first stage of PGMT learns a general whole-body motion tracking prior on flat terrain, providing a stable initialization for the subsequent injection of terrain perception. This stage adopts a terrain-agnostic policy architecture mainly consisting of an Intent Fusion Module (IFM), an actor, and a multi-head critic. The IFM integrates proprioceptions and future reference frames into a state-conditioned representation, which is used by the actor for low-level motion control and by the critic for value estimation. To train this stage, we combine a body-part-based reward decomposition with a recovery curriculum and adaptive motion sampling, enabling a unified policy to learn diverse whole-body motions.

\subsubsection{Terrain-agnostic Policy Architecture}
\label{sec:pretrain_architecture}

Given the current proprioceptive observation $\bm{o}_t$ and the proprioceptive history $\bm{H}_t$, we use $\bm{o}_t$ as the query to attend to the historical observations in $\bm{H}_t$, producing a compact history-conditioned state token:
\begin{equation}
\bm{s}_t^{\mathrm{hist}}
=
\operatorname{CrossAttn}_{H}
\left(
Q=\bm{o}_t,\,
K=V=\bm{H}_t
\right),
\label{eq:pretrain_history}
\end{equation}
where $\operatorname{CrossAttn}_{\cdot}$ denotes a multi-head cross-attention (MHCA) layer, and $Q$, $K$, and $V$ denote the query, key, and value, respectively. The resulting token $\bm{s}_t^{\mathrm{hist}}$ provides a compact representation of recent robot dynamics and temporal context that is not available from the instantaneous proprioceptive observation alone.

\noindent\textbf{Intent Fusion Module.}
To effectively fuse proprioceptive information with future motion references, we design the IFM consisting of an MHCA layer equipped with Rotary Position Embedding (RoPE). Specifically, the module uses the history-conditioned state token $\bm{s}_t^{\mathrm{hist}}$ as the query to attend to the concatenation of the current state token and future reference frames $\bm{C}_t^{K}$, producing a state-conditioned motion-intent token:
\begin{equation}
\bm{s}_t^{\mathrm{int}}
\!=\!
\operatorname{CrossAttn}_{C}
\left(
Q\!=\!\bm{s}_t^{\mathrm{hist}},\,
K\!=V\!=[\bm{s}_t^{\mathrm{hist}};\bm{C}_t^{K}]
\right).
\label{eq:pretrain_intent}
\end{equation}
This design allows the policy to selectively emphasize future reference information conditioned on the robot's current state and recent dynamics. 
The resulting intent token $\bm{s}_t^{\mathrm{int}}$ is shared by the actor and critic and is further augmented with terrain information in the second training stage (Sec.~\ref{sec:terrain_extension}).

\noindent\textbf{Actor Network.}
The actor concatenates the current proprioceptive $\bm{o}_t$ with the motion-intent token $\bm{s}_t^{\mathrm{int}}$ and maps the resulting representation to the control action:
\begin{equation}
\bm{a}_t
=
\operatorname{MLP}_{A}
\left(
[\bm{o}_t,\bm{s}_t^{\mathrm{int}}]
\right),
\label{eq:pretrain_actor}
\end{equation}
where $\operatorname{MLP}$ is a multi-layer perceptron. While $\bm{s}_t^{\mathrm{int}}$ provides a compact representation of the state-conditioned motion intent, the direct inclusion of
$\bm{o}_t$ preserves instantaneous proprioceptive feedback for
fine-grained low-level control. The target joint positions are
converted into joint torques by a low-level PD controller.
The actor relies exclusively on information available at deployment
time.

\noindent\textbf{Multi-Head Critic.} Whole-body motion tracking combines objectives with distinct learning characteristics. Upper-body objectives primarily preserve motion expression and pose fidelity, whereas lower-body objectives are closely coupled with balance and contact transitions. The remaining objectives shape root motion, recovery, safety, and control regularization. As summarized in Table~\ref{tab:reward_groups}, we partition the pretraining reward into three groups:
\begin{equation}
r_t = r_t^{\mathrm{upper}} + r_t^{\mathrm{lower}} + r_t^{\mathrm{aux}},
\end{equation}
where $r_t^{\mathrm{upper}}$ and $r_t^{\mathrm{lower}}$ contain the upper- and lower-body tracking objectives, respectively, and $r_t^{\mathrm{aux}}$ contains the remaining auxiliary objectives. A single scalar critic may entangle reward components with substantially different scales and learning dynamics. We therefore employ a multi-head critic that outputs a vector of value estimates $\bm{V}_t$, one per reward group:
\begin{equation}
\bm{V}_t = \operatorname{MLP}_{V}\left([\bm{o}_t^{\mathrm{priv}}, \bm{s}_t^{\mathrm{int}}]\right) = \left[V_t^{\mathrm{upper}}, V_t^{\mathrm{lower}}, V_t^{\mathrm{aux}}\right],
\label{eq:pretrain_critic}
\end{equation}
where $\bm{o}_t^{\mathrm{priv}}$ denotes privileged observations available only during simulation training. The three heads share a common critic backbone, enabling representation sharing while modeling the reward groups separately. 

\subsubsection{Training Optimization}
\label{sec:pretrain_setup}

In this stage, we train PGMT using motions from LAFAN1~\cite{harvey2020robust} retargeted to a humanoid robot. Motion segments are sampled across reference sequences and starting frames to train a unified policy over diverse whole-body behaviors. All pretraining is conducted on flat terrain.
We adopt a root-centric tracking formulation, where the primary body-tracking errors are computed relative to the reference root anchor. In contrast to tracking full trajectories in the world frame, this formulation emphasizes relative body motion and reduces dependence on absolute global position and heading. Consequently, the learned tracking prior can reproduce the same motion intent from different initial poses and headings, providing a transferable initialization for subsequent terrain adaptation.

\noindent\textbf{Recovery Curriculum and Adaptive Sampling.}
Following RGMT~\cite{ma2026robust}, we incorporate the recovery curriculum into this stage's training. We construct an offline pool of post-fall robot states and progressively increase the proportion of environments initialized from this pool according to episode-survival performance. In parallel, we employ adaptive sampling that assigns higher probabilities to motion segments associated with frequent tracking failures while retaining uniform coverage of the full motion dataset.


\begin{table}[t]
\centering
\caption{Reward groups for PGMT.}
\label{tab:reward_groups}
\setlength{\tabcolsep}{2.5pt}
\renewcommand{\arraystretch}{0.98}
\resizebox{\columnwidth}{!}{%
\begin{tabular}{@{}lrlr@{}}
\toprule
Reward term & Weight & Reward term & Weight \\
\specialrule{\lightrulewidth}{\aboverulesep}{0pt}
\rowcolor{gray!15}
\multicolumn{4}{@{}c@{}}{\emph{The 1st \& 2nd Stages}} \\
\multicolumn{4}{@{}l}{\textbf{Upper body}} \\
Link position
    & $1.0$
    & Link orientation
    & $1.0$ \\
Link linear velocity
    & $0.5$
    & Link angular velocity
    & $0.5$ \\
Joint position
    & $1.0$
    & Joint velocity
    & $0.5$ \\

\addlinespace[1pt]
\multicolumn{4}{@{}l}{\textbf{Lower body}} \\
TA link position
    & $0.5$
    & TA link orientation
    & $2.0$ \\
Link linear velocity
    & $0.5$
    & Link angular velocity
    & $0.5$ \\
TA joint position
    & $0.5$
    & Joint velocity
    & $0.5$ \\

\addlinespace[1pt]
\multicolumn{4}{@{}l}{\textbf{Auxiliary}} \\
Root orientation
    & $0.5$
    & Corrected root velocity
    & $2.0$ \\
Floating-anchor position
    & $1.0$
    & Recovery upward velocity
    & $12.5$ \\
Pelvis vertical acceleration
    & $-10^{-3}$
    & EE acceleration mismatch
    & $-10^{-3}$ \\
Action rate
    & $-0.05$
    & Joint limit
    & $-15.0$ \\
Undesired contact
    & $-0.1$
    & Head/torso impact
    & $-10^{-5}$ \\

\addlinespace[1pt]
\rowcolor{gray!15}
\multicolumn{4}{@{}c@{}}{\emph{The 2nd Stage}} \\
\multicolumn{4}{@{}l}{\textbf{Terrain-contact}} \\
Touchdown quality
    & $10.0$
    & Reference contact match
    & $1.5$ \\
Slip
    & $-1.0$
    & Stumble
    & $-20.0$ \\
Contact switching
    & $-30.0$
    & Contact force
    & $-10^{-6}$ \\

\bottomrule
\end{tabular}%
}
\vspace{-0.8cm}
\end{table}

\subsection{Perception Injection Stage}
\label{sec:terrain_extension}

At this stage, we progressively extend the PGMT policy architecture through a perception injection process, enabling PGMT to adapt to diverse terrains. 


\subsubsection{Perception Injection Process}

Building upon the terrain-agnostic policy architecture, we introduce a Terrain-Glimpse Encoder that generates state-conditioned terrain tokens, which are subsequently integrated into the IFM to facilitate terrain-aware low-level motion control. In addition, we extend the multi-head critic to evaluate a newly introduced terrain-contact objective.

\noindent\textbf{Terrain-glimpse Encoder.} Directly encoding the entire elevation map as a dense terrain representation for the actor is computationally expensive and may introduce unnecessary noise, as only a small subset of terrain regions is typically relevant to locomotion, particularly those around potential footholds and terrain boundaries. Instead, PGMT employs a Terrain-Glimpse Encoder that leverages the elevation map to localize a sparse set of locomotion-relevant regions conditioned on the robot’s dynamics and anticipated motions. These regions are then selectively sampled and encoded to provide compact terrain information for low-level motion control.
Given the history-conditioned state token $s_t^\mathrm{hist}$, the elevation map $M_t$, and the future reference frames $C_t^K$, the encoder first employs a position selector, implemented as an MLP, to predict $N_g=4$ glimpse locations:
\begin{equation}
\bm{r}_t^j\! =\! \left[ \operatorname{MLP}_{\phi}\bigl( \operatorname{vec}(\bm{M}_t),\bm{s}_t^{\mathrm{hist}},\bm{C}_t^K\bigr)\right]_j, j \!\in \!\{1,\! \cdots,\!N_g\},
\end{equation}
where $\bm{r}_t^j\in \mathbb{R}^2$ denotes the \(j\)-th predicted sampling location and $\operatorname{vec}(\cdot)$ denotes the vectorization operation that flattens its input into a one-dimensional vector. Then, we crop $N_g$ terrain patches of size $5\times5$ centered at the predicted locations $\bm{r}_t^j$. Each patch, together with its center position $\bm{\rho}_t^j$ expressed in the robot frame, is passed through an MLP to obtain the local terrain token:
\begin{equation}
\bm{z}_t^j\! =\!
\operatorname{MLP}_{\psi}\left(
\operatorname{Crop}_{5\times5}(\bm{M}_t,\bm{r}_t^j),
\bm{\rho}_t^j
\right), j\!=\!\{1,\!\cdots,\!N_g\}.
\label{eq:terrain_glimpses}
\end{equation}
Bilinear sampling makes the cropping operation differentiable,
allowing the glimpse locations to be learned without region or
foothold annotations.
The resulting local terrain tokens are passed to the IFM.

\noindent\textbf{Integration with IFM.} In the first stage, the IFM produces the motion-intent token $\bm{s}_t^{\mathrm{int}}$ conditioned solely on proprioceptive information and future motion references. At this stage, we inject terrain information into the IFM by augmenting the keys and values in Eq.~(\ref{eq:pretrain_intent}) with the local terrain tokens $\{\bm{z}_t^j\}_{j=1}^{N_g}$: $K=V=\left[\bm{s}_t^{\mathrm{hist}};\bm{C}_t^K;\bm{z}_t^1;\ldots;\bm{z}_t^{N_g}\right]$.
This enables the motion-intent token to incorporate terrain-aware information for low-level motion control.

\noindent\textbf{Multi-Head Critic Extension.}
To encourage stable contacts, we introduce a new terrain-contact reward group $r_t^{\mathrm{terrain}}$ (summarized in Table~\ref{tab:reward_groups}). Local height variation is used to evaluate touchdown quality, while contact labels obtained from offline terrain-mesh queries supervise the consistency between reference and simulated contacts. Additional penalties discourage foot slippage, stumbling, rapid contact switching, and excessive contact forces.
To accommodate the new reward group $r_t^{\mathrm{terrain}}$, we extend the multi-head critic from the first stage with an additional value head $V_t^{\mathrm{terrain}}$ that estimates the expected terrain-contact return. This is achieved by modifying the final layer of $\operatorname{MLP}_V$ to output four values instead of three. The values estimated by the multi-head critic in this stage are then given by
\begin{equation}
\bm{V}_t^{\mathrm{inj}}=\left[V_t^{\mathrm{upper}},V_t^{\mathrm{lower}},V_t^{\mathrm{terrain}},V_t^{\mathrm{aux}}\right].
\label{eq:injection_critic}
\end{equation}
This dedicated value head provides a focused learning signal for terrain interaction and foothold selection, without interfering with the value estimates of the other reward groups.


\subsubsection{Terrain-Adaptive Training Design}

Beyond the architectural extension, the second stage requires several training-level adjustments so that the policy can deviate from the flat-ground reference where the terrain demands it. 

\noindent\textbf{Terrain-Aware Tracking Relaxation.}
Strictly penalizing every tracking error would suppress necessary changes in footholds, swing trajectories, and lower-body posture. We therefore introduce a terrain-aware relaxation into selected tracking objectives: $\widehat e_{t,h,j}=\left[e_{t,h,j}-\alpha_{h,j}\,\chi(\kappa_t)\,\tau_{m_h}(d_t)\right]_+ $,
where $e_{t,h,j}$ is the raw error of tracking objective $h$ at link or joint $j$, $[\cdot]_+ = \max(0,\cdot)$, and $\widehat e_{t,h,j}$ replaces $e_{t,h,j}$ in the exponential tracking reward. Here, $\kappa_t$ and $d_t$ denote the terrain family and difficulty level of the tile under the robot: the indicator $\chi(\kappa_t)\in\{0,1\}$ activates the relaxation only on slopes, stairs, and boxes; $\tau_{m_h}(d_t)$ is a base relaxation in the unit of objective $h$ that grows linearly with $d_t$ and saturates; and $\alpha_{h,j}\ge 0$ scales it per element, with $\alpha_{h,j}=0$ recovering strict tracking. The relaxation is applied only to lower-body link position, link orientation, and joint position objectives, while upper-body tracking objectives and remaining motion constraints remain strict, preserving the reference motion while permitting necessary lower-body adaptation.

\noindent\textbf{Global Position Correction.}
To prevent accumulated global position error, we feed the planar position error back into the reference velocity command. The error is computed in the reference anchor frame as $\bm{e}_{t,xy}^{p}=\left[(\bm{R}_t^{r,w})^\top(\bm{p}_t^{r,w}-\bm{p}_t^w)\right]_{xy}$, where $\bm{p}_t^{r,w}$ and $\bm{p}_t^{w}$ are the world-frame anchor positions of the reference and the robot, $\bm{R}_t^{r,w}$ is the world-frame rotation of the reference anchor, and $[\cdot]_{xy}$ extracts the planar components. The nominal planar reference velocity $\bm{v}_{t,xy}^{r}$ is then corrected as $\widetilde{\bm{v}}_{t,xy}^{r}=\bm{v}_{t,xy}^{r}+\operatorname{clip}_{[-\bar v,\bar v]}\!\left(g(\|\bm{v}_{t,xy}^{r}\|_2)\,\lambda_{\mathrm{pos}}\,\bm{e}_{t,xy}^{p}\right)$, i.e., by a proportional term with gain $\lambda_{\mathrm{pos}}$, modulated by a saturating speed gate $g(\cdot)\ge 0$ that grows with the nominal speed and vanishes for a stationary reference, and clipped component-wise at $\pm\bar v$. The corrected velocity serves as both the policy command and the velocity-tracking target. Setting $\lambda_{\mathrm{pos}}=0$ recovers pure velocity tracking, whereas $\lambda_{\mathrm{pos}}>0$ enables global position correction.


\noindent\textbf{Terrain Curriculum and Reference Sampling.} We train on flat terrain, slopes, stairs, boxes, and random rough terrain using a level-based curriculum, in which environments advance to harder terrain after successful completion and regress to easier levels after tracking failures. The terrain level jointly controls geometric difficulty, tracking relaxation, and selected termination delays. Since not every motion is feasible on every terrain, we use a coarse behavior--terrain compatibility rule to exclude clearly invalid combinations when sampling references. This requires neither terrain-adapted references, explicit foothold targets, nor paired motion--terrain demonstrations.

\begin{table*}[t]
    \centering
\caption{\textbf{Evaluation performance and observed training cost.}
Completion denotes reaching the 30-s horizon without early termination;
Level 9 (L9) reports completion over 960 highest-difficulty episodes.}
\label{tab:main_results}
\scriptsize
\setlength{\tabcolsep}{2.5pt}
\resizebox{\textwidth}{!}{%
\begin{tabular}{lrrrrrrrrr}
    \toprule
    & \multicolumn{6}{c}{Evaluation}
    & \multicolumn{3}{c}{Observed training cost} \\
    \cmidrule(lr){2-7}\cmidrule(lr){8-10}
    Policy
    & \shortstack{Completion\\(\%) $\uparrow$}
    & \shortstack{L9\\(\%) $\uparrow$}
    & \shortstack{Body Pos.\\(m) $\downarrow$}
    & \shortstack{Body Ori.\\($^\circ$) $\downarrow$}
    & \shortstack{Joint RMSE\\(rad) $\downarrow$}
    & \shortstack{Contact\\F1 $\uparrow$}
    & \shortstack{Time\\(s/iter.) }
    & \shortstack{Max VRAM\\(GiB) }
    & \shortstack{Envs/\\GPU} \\
    \midrule
    \multicolumn{10}{l}{\emph{Flat-only baselines on terrains}} \\
    PGMT-Pretrain
    & 40.02
    & 35.31
    & 0.0897
    & 27.70
    & 0.2957
    & \textbf{0.7436}
    & 4.54
    & 16.11
    & 10k \\
    RGMT-Reimpl~\cite{ma2026robust}
    & 46.68 
    & 40.52 
    & 0.0700 
    & 20.15 
    & \textbf{0.2251} 
    & 0.7680
    & 5.42 
    & 24.78 
    & 8{,}192 \\
    SONIC v1.1 External$^{\dagger}$~\cite{luo2026sonic}
    & 25.09
    & 20.73
    & 0.1204
    & 36.42
    & 0.2650
    & 0.8232
    & --
    & --
    & --\\

    \midrule
    \multicolumn{10}{l}{\emph{Comparisons trained on terrains}} \\
    Perceptive BFM (paper-reported)$^{*}$~\cite{wang2026perceptive}
    & 55.1
    & --
    & --
    & --
    & --
    & --
    & --
    & --
    & -- \\

    PGMT-NoHeight
    & 86.53
    & 78.85
    & 0.0659
    & 20.32
    & 0.2937
    & 0.8157
    & 8.47
    & 29.88
    & 15k \\
    PGMT-CNN
    & 87.38
    & 80.52
    & \textbf{0.0645}
    & 18.33
    & 0.2397
    & 0.8173
    & 11.95
    & 27.30
    & 15k \\
    PGMT-2Glimpse
    & 85.17
    & 77.71
    & 0.0676
    & 18.27
    & 0.2340
    & 0.8234
    & 8.36
    & 28.10
    & 15k \\
    PGMT-Fixed
    & 86.38
    & 79.58
    & 0.0645
    & 18.47
    & 0.2461
    & 0.8218
    & 8.28
    & 28.66
    & 15k \\
    \midrule
    \textbf{PGMT-Ours}
    & \textbf{87.81}
    & \textbf{83.33}
    & 0.0678
    & \textbf{18.14}
    & 0.2299
    & 0.8217
    & 8.44
    & 29.23
    & 15k \\
    \bottomrule
\end{tabular}%
}

\vspace{2pt}
\begin{minipage}{0.99\linewidth}
\footnotesize
RMSE denotes root-mean-square error.
Training costs are reported only for runs trained in this work using
four RTX~5090 GPUs; Time is the mean wall-clock time per PPO iteration,
and VRAM is the peak single-GPU usage.
$^{\dagger}$External-checkpoint results use re-audited adapters and
checkpoint-native controllers.
$^{*}$Perceptive BFM results are quoted from the original work.
\end{minipage}
\vspace{-10pt}
\end{table*}

\section{Experiments}
\label{sec}


We systematically evaluate PGMT in simulation and real-world deployment. In simulation, we compare it with general motion-tracking baselines and alternative terrain-perception designs across terrain types and difficulty levels. We then test it across multiple real-world control modes.

\subsection{Experimental Setup}


We evaluate terrain-adaptive motion tracking on five terrain families:
flat terrain, slopes, stairs, box obstacles, and randomly rough terrain.
Each family contains ten difficulty levels, denoted by L0--L9 in
increasing order of difficulty. Each policy is evaluated over 9,600
matched 30-s episodes, with 192 episodes for each terrain--level pair.
Corresponding episodes use identical motion references, starting
frames, terrain assignments, dynamics randomization, observation
corruption, and actuator-delay settings. An episode is considered
successful if it reaches the maximum rollout horizon without early
termination. Results are reported in Table~\ref{tab:main_results}.

\begin{figure}
\centering
\includegraphics[width=1\linewidth,trim=0 0 0 100,clip]{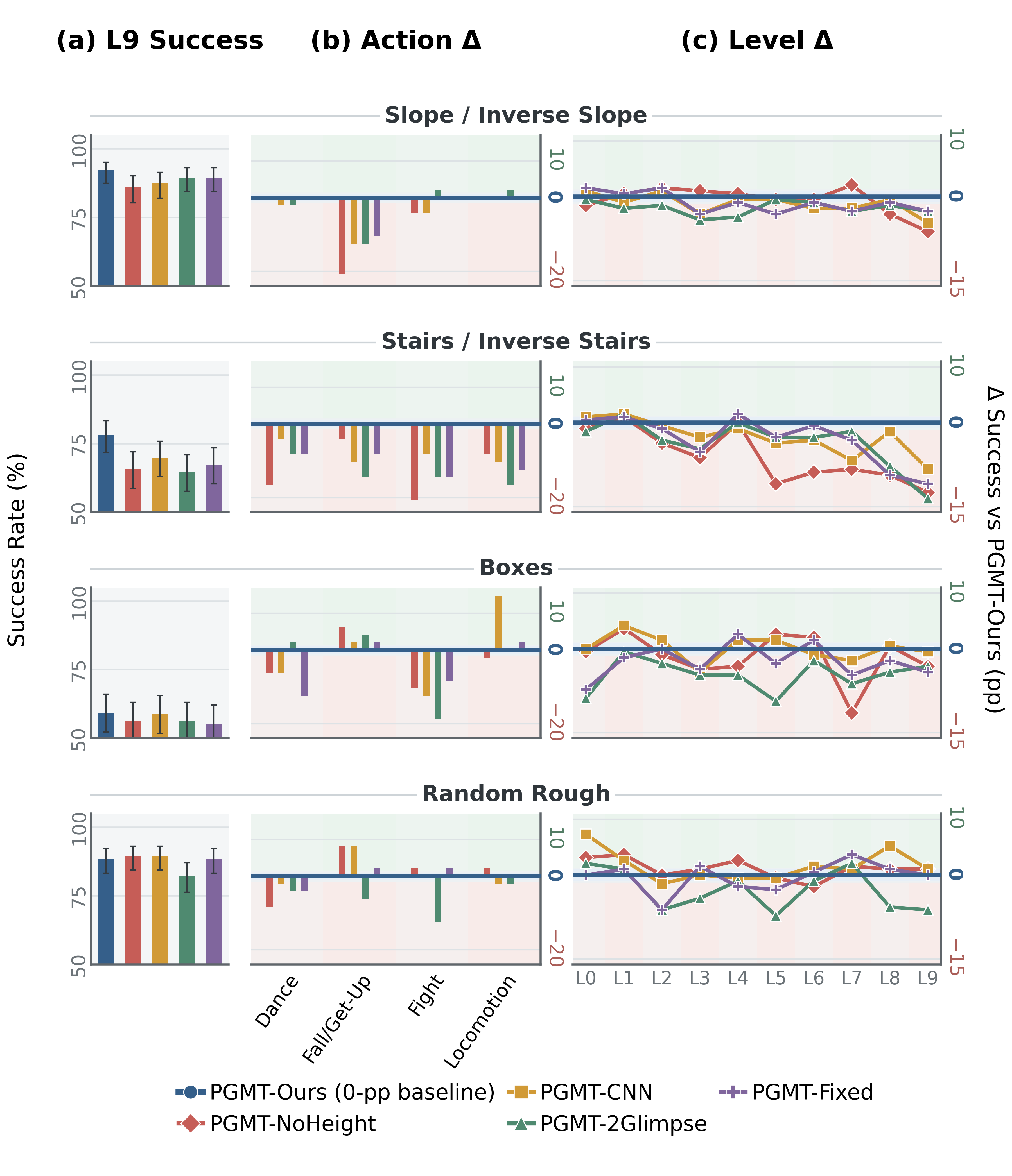}
\caption{\textbf{Terrain-perception ablations.}
(a) L9 success with Wilson 95\% intervals.
(b) L9 motion-group success differences from PGMT-Ours (48 episodes per bar).
(c) Success differences from PGMT-Ours across L0--L9.}
\vspace{-0.8cm}
\label{fig:ablation}
\end{figure}

\begin{figure}
    \centering
    \includegraphics[width=\linewidth]{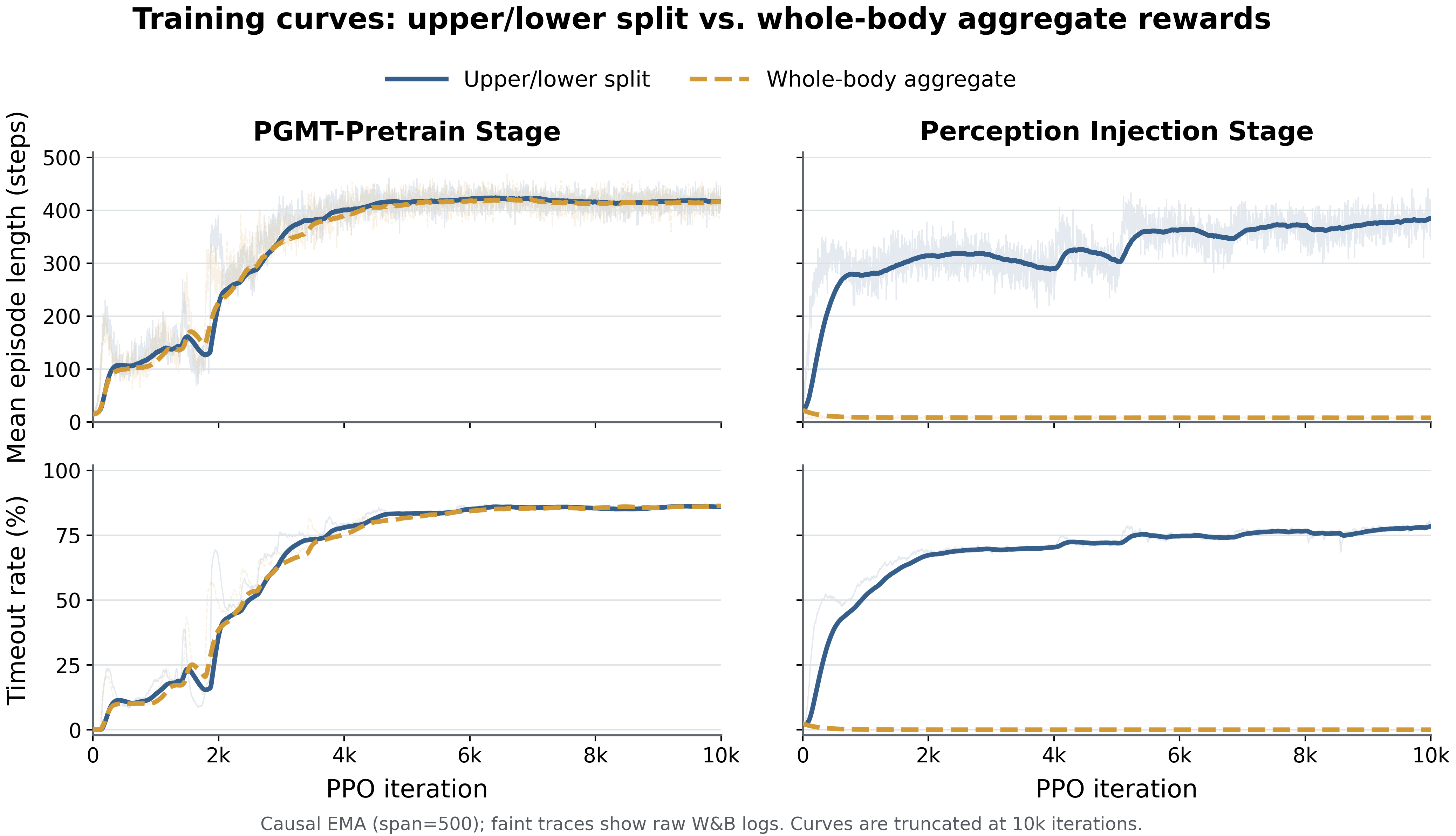}
\caption{\textbf{Split- and aggregate-return critics.}
Both learn during Tracking Pretraining, but only the split-return critic progresses during Perception Injection.
Curves show causal EMAs of the raw logs, truncated at 10k PPO iterations.}
    \label{fig:critic_ablation}
    \vspace{-0.8cm}
\end{figure}

\begin{figure*}
    \centering
    \includegraphics[width=0.95\linewidth]{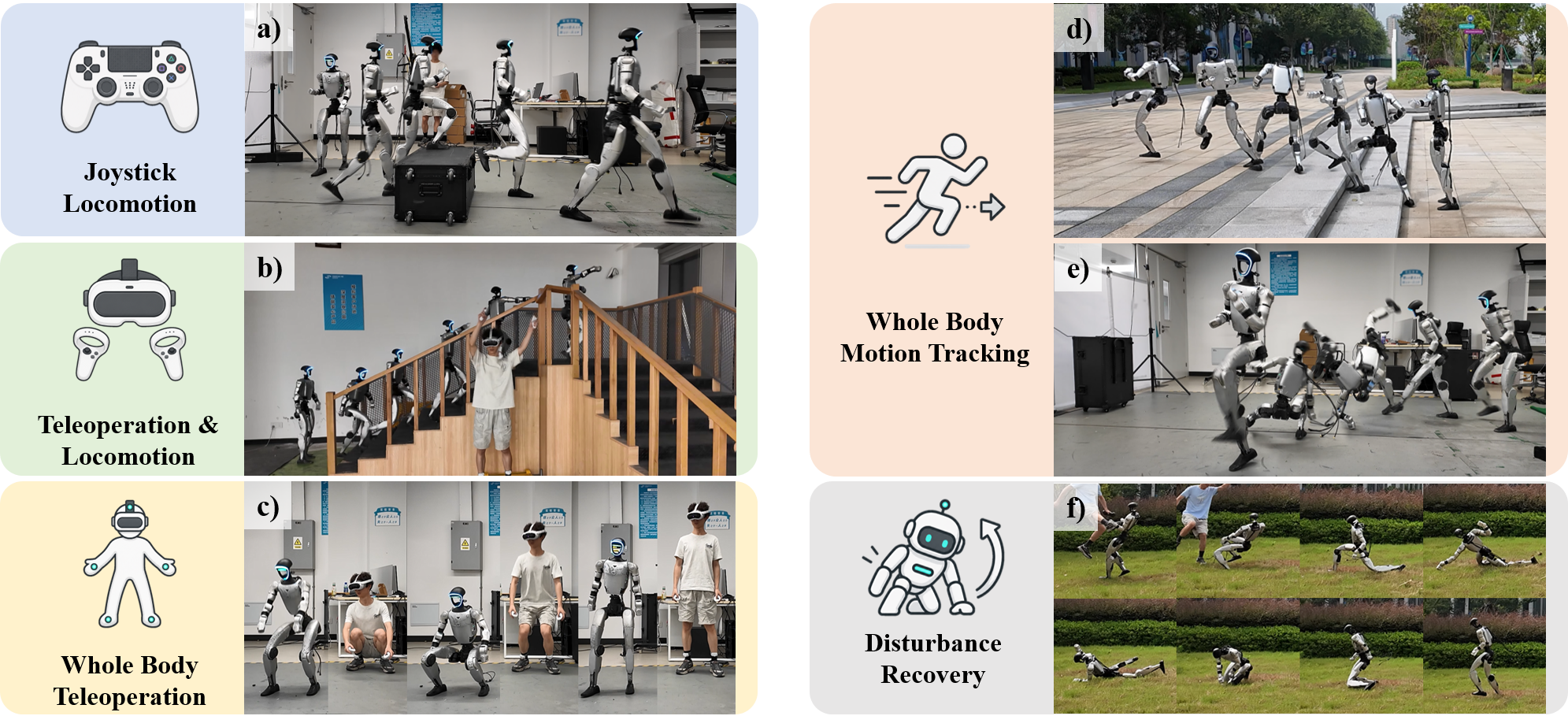}
    \caption{\textbf{Unified real-world control with PGMT.} A single unified PGMT policy supports diverse command sources and control modes, including \textbf{(a)} joystick locomotion, \textbf{(b)} teleoperated locomotion, \textbf{(c)} whole-body teleoperation, \textbf{(d-e)} whole-body motion tracking, and \textbf{(f)} disturbance recovery. Locomotion commands are given through Unitree or PICO joystick. All behaviors are executed by the same policy without policy switching.}
    \label{fig:control_mode}
    \vspace{-0.8cm}
\end{figure*}

\subsection{Comparison with Motion-Tracking Baselines}

We compare PGMT with several general motion-tracking baselines, including \textbf{PGMT-Pretrain} before Perception Injection, our reimplementation \textbf{RGMT-Reimpl}~\cite{ma2026robust}, and the external checkpoint \textbf{SONIC v1.1 External}~\cite{luo2026sonic}. We also include the published Perceptive BFM result~\cite{wang2026perceptive}.

As shown in Table~\ref{tab:main_results}, PGMT-Ours achieves completion
rates of 87.81\% over the full benchmark and 83.33\% at L9, whereas
PGMT-Pretrain reaches only 40.02\% and 35.31\%, respectively. The
other motion trackers without terrain observations exhibit similar
performance degradation. These results show that a general motion
prior alone cannot reliably resolve mismatches between reference
motions and local terrain, while terrain perception substantially
improves tracking success on complex terrains.

\subsection{Ablation Studies}

We first evaluate four variants that modify only the terrain-perception mechanism. \textbf{PGMT-NoHeight} forces a zero input for terrain perception. \textbf{PGMT-CNN} replaces the glimpse module with a CNN that encodes the complete height scan into a fixed $2\times2$ terrain-token grid. \textbf{PGMT-2Glimpse} reduces the number of terrain glimpses from four to two. \textbf{PGMT-Fixed} uses four predefined glimpse locations instead of dynamically predicting them from the robot history and upcoming motion. All other policy components and training settings remain unchanged.

As shown in Table~\ref{tab:main_results}, PGMT-Ours achieves the highest overall completion rate (87.81\%) and L9 completion rate (83.33\%). Across the four non-flat terrain families at L9 in Figure~\ref{fig:ablation}, PGMT-Ours achieves an average success rate of 79.56\%, compared with 76.43\% for the strongest alternative, PGMT-CNN. The largest advantage occurs on stairs, where PGMT-Ours achieves 78.12\%, whereas the ablated variants achieve only 64.58--69.79\%. Over the full evaluation schedule, PGMT-CNN remains comparable to PGMT-Ours. However, the larger separation on L9 stairs indicates that motion-conditioned spatial selection is particularly beneficial when complex terrain geometry and difficult motions occur together. These results further show that our adaptive region selection improve robustness on challenging terrains.

We further evaluate the effectiveness of the proposed multi-head split-return critic by replacing it with a single-head aggregate-return critic that estimates a single whole-body return. The split-return critic uses a shared backbone and separate heads to estimate the returns associated with the upper-body, lower-body, and auxiliary reward groups. As shown in Figure~\ref{fig:critic_ablation}, both critics successfully learn the terrain-agnostic motion-tracking policy during the Tracking Pretraining Stage, achieving comparable mean episode lengths and timeout rates. Once terrain perception is introduced, however, their training behaviors diverge substantially. The split-return critic continues to make sustained progress, eventually reaching a mean episode length of approximately 380 steps and a timeout rate of nearly 80\%. By contrast, both metrics remain near zero for the aggregate-return critic.

\subsection{Real-World Evaluation}

We deploy the PGMT policy zero-shot on a 29-DoF Unitree G1. A Livox Mid-360S LiDAR and an onboard elevation-mapping pipeline provide terrain observations matching the simulation format. Perception and control run on a single NVIDIA Jetson Orin NX. Representative executions are shown in Fig.~\ref{fig:frontpage}, and Fig.~\ref{fig:control_mode} summarizes the supported control modes.


\subsubsection{Terrain-Adaptive Locomotion}

We evaluate terrain-adaptive locomotion in both indoor and outdoor environments using flat-ground motion references generated online through motion matching. PGMT traverses uneven ground, ascends and descends staircases with unseen geometries, and climbs onto boxes up to $37\,\mathrm{cm}$ high. Across these terrains, the policy adapts foot placement, swing clearance, and body posture according to the local elevation map while following the locomotion intent.

We further evaluate the policy on irregular outdoor terrain, including grass and densely vegetated surfaces, as well as during jogging and sprinting on stairs, where the robot can clear multiple steps in a single stride (Fig.~\ref{fig:frontpage}). Despite soft support surfaces and substantial perception noise, PGMT maintains stable locomotion without real-world fine-tuning.

\subsubsection{Teleoperation}

We demonstrate two teleoperation modes with the same PGMT policy: whole-body teleoperation and locomotion teleoperation. In whole-body teleoperation, the robot executes diverse operator-commanded behaviors, including, punching, squatting, lying down, and recovering to a standing posture. The operator's full-body motion is captured through PICO and retargeted online to the robot using GMR (Fig.~\ref{fig:control_mode}c).

In locomotion teleoperation, the operator controls the robot's locomotion together with upper-body motion, while the lower-body reference is generated online through motion matching. As shown in Fig.~\ref{fig:control_mode}b, the robot can climb a flight of stairs while the operator simultaneously teleoperates the upper body. PGMT adapts the locomotion reference to the local terrain, allowing the operator to control the upper body during terrain traversal without explicitly specifying terrain-dependent lower-body motions.

\subsubsection{General Dynamic Whole-Body Motion Tracking}

Beyond locomotion and teleoperation, we evaluate PGMT on diverse whole-body motions in the real world. PGMT retains the ability to execute highly dynamic motions (Fig.~\ref{fig:control_mode}e) including cartwheels. We further evaluate terrain aware motion tracking non-flat terrain (Fig.~\ref{fig:control_mode}d). PGMT executes these whole-body references on stairs and other uneven surfaces while adapting its contacts and posture to the local terrain. This demonstrates that terrain adaptation does not restrict PGMT to locomotion: the policy can preserve the structure of diverse whole-body motions while making the terrain-dependent adjustments required for stable execution.

\subsubsection{Disturbance Robustness and Fall Recovery}

Finally, we evaluate PGMT under large external disturbances (Fig.~\ref{fig:control_mode}f). The robot withstands strong pushes and kicks while performing a dance, and recovers its posture after substantial perturbations without interrupting the commanded motion.

When a sufficiently large disturbance causes a fall, PGMT autonomously recovers from the fallen configuration and returns to a stable standing posture. Both disturbance rejection and fall recovery are handled by the same unified policy without switching to a dedicated recovery controller, demonstrating robust whole-body control across both upright and fallen states.

\section{Conclusion}

We presented PGMT, a perceptive general motion tracker that adapts terrain-agnostic reference motions to complex terrain without requiring terrain-matched demonstrations or explicitly generated reference trajectories. By injecting motion-conditioned terrain perception into a general tracking prior, PGMT extends general whole-body motion tracking beyond flat ground. Beyond general motion tracking, the same policy can serve directly as a terrain-adaptive locomotion controller or a whole-body teleoperation interface, and more broadly as a perceptive low-level controller that grounds terrain-agnostic motion commands from higher-level systems into physically feasible behaviors. 
A current limitation is that elevation maps encode geometry but not environmental semantics or affordances, making it difficult to distinguish obstacles from objects intended for interaction. Incorporating richer perceptual representations could further extend PGMT from terrain-aware motion execution toward more general humanoid interaction in unstructured environments.









\bibliographystyle{IEEEtran}
\bibliography{references}

\end{document}